\pdfoutput=1

\documentclass[11pt]{article}

\usepackage{EMNLP2022}

\usepackage{times}
\usepackage{latexsym}
\usepackage{graphicx}

\usepackage[T1]{fontenc}

\usepackage[utf8]{inputenc}

\usepackage{microtype}

\usepackage{inconsolata}

\usepackage{comment}

\usepackage{booktabs}

\usepackage{amsmath}
\usepackage{amssymb}
\usepackage{bm} 
\usepackage{dsfont}

\usepackage{watermark}
\usepackage[calc,useregional]{datetime2}

\usepackage{enumitem}
\usepackage{defs}

\DeclareMathOperator*{\argmax}{argmax}

\title{Predicting Long-Term Citations from Short-Term Linguistic Influence}

\author{Sandeep Soni \and David Bamman \\
  University of California, Berkeley \\
  \texttt{\{sandeepsoni,dbamman\}@berkeley.edu} \\\And
  Jacob Eisenstein \\
  Google Research\\
  \texttt{jeisenstein@google.com}\\
  }

\begin{document}
\maketitle
\begin{abstract}
A standard measure of the influence of a research paper is the number of times it is cited. However, papers may be cited for many reasons, and citation count offers limited information about the extent to which a paper affected the content of subsequent publications. We therefore propose a novel method to quantify linguistic influence in timestamped document collections. There are two main steps: first, identify lexical and semantic changes using contextual embeddings and word frequencies; second, aggregate information about these changes into per-document influence scores by estimating a high-dimensional Hawkes process with a low-rank parameter matrix. We show that this measure of linguistic influence is predictive of \emph{future} citations: the estimate of linguistic influence from the two years after a paper's publication is correlated with and predictive of its citation count in the following three years. This is demonstrated using an online evaluation with incremental temporal training/test splits, in comparison with a strong baseline that includes predictors for initial citation counts, topics, and lexical features.

\end{abstract}
\section{Introduction}
The citation count of a paper is a standard, easily measurable proxy for its influence~\citep{cronin2005hand}. Researchers have shown that citation count is strongly correlated with the quality of scientific work~\citep[e.g.,][]{lawani1986some}, the recognition that a paper or an author gets~\citep[e.g.,][]{inhaber1976quality}, or in policy decisions such as assessment of scientific performance~\citep[e.g.,][]{cronin2005hand}. Consequently, citation count is a ubiquitously deployed and important measure of a paper with whole subfields of research dedicated to its analysis~\citep{bornmann2008citation}. 

However, papers may be cited (or not cited) for many reasons, and citation count alone is insufficient to explain the emergence and the spread of research ideas and trends. For this reason, we turn to content analysis: to what extent can the text of a research paper be said to influence the trajectory of the research community? In this paper, we present a novel technique for estimating the influence of documents in a timestamped corpus. To demonstrate the validity of the resulting measure of linguistic influence, we show that it is predictive of \emph{future} citations. Specifically, we find that: (1) papers that our metric judges as highly influential in the short term tend to receive more citations in the long term; (2) short-term linguistic influence increases the ability to predict long-term citations over strong baselines.

\begin{figure}
    \centering
    \includegraphics[width=\linewidth]{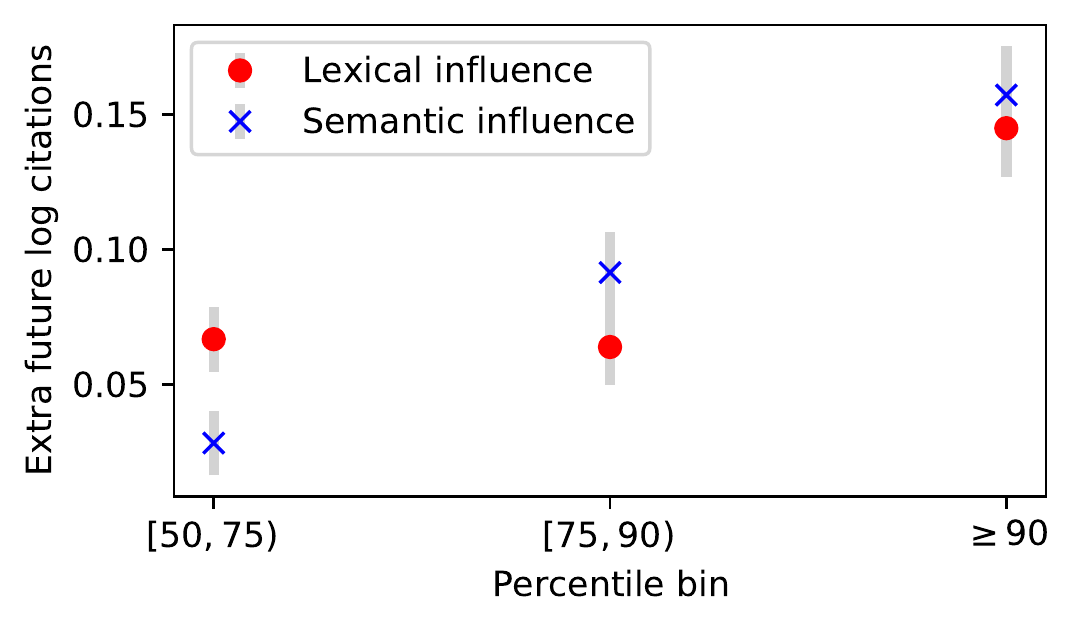}
    \caption{Research papers that are more linguistically influential within an initial time window tend to receive more citations in the long term. The $x$-axis shows lexical and semantic influence, binned into quantiles (see \autoref{sec:methodology}); the $y$-axis shows the corresponding regression coefficients and standard errors, in units of $Z$-normalized log future citations (see \autoref{sec:results-regression}). To give a sense of scale, 
    for papers published in 2012, being in the top decile of semantic influence corresponds to an $14.5$\% increase in long-term citations, as compared to control-matched papers that received the same number of short-term citations and covered similar topics but were in the bottom half by semantic influence.
    }
    \label{fig:quantile-plot}
\end{figure}

Our modeling approach focuses on semantic changes, and treats the temporal usage of semantic innovations as emissions from a parametric low-rank Hawkes process~\citep{hawkes1971spectra}. The parameters of the Hawkes process correspond to the linguistic influence of each paper, aggregated over thousands of linguistic changes. The changes themselves are identified through analysis of contextual embeddings, with the goal of finding words whose meaning has shifted over time~\citep{traugott2001regularity}. Though there are several computational methods to detect semantic changes~\citep[e.g.,][]{kim-etal-2014-temporal, hamilton-etal-2016-diachronic, rosenfeld-erk-2018-deep, dubossarsky-etal-2019-time}, including methods based on contextual embeddings~\citep[e.g.,][]{kutuzov-giulianelli-2020-uio}, our proposed method focuses on detecting smooth, non-bursty semantic changes; we also go further than other methods by distinguishing old and contemporary usages of an identified semantic change.

We show through a multivariate regression that our estimates of semantic influence of each paper are positively correlated with their long-term citations, even after controlling for the initial citations, the content of the paper in terms of topics, and the lexical influence of the paper  (see~\autoref{fig:quantile-plot}). Further, we formulate long-term citation prediction as an online prediction task, constructing test sets for successive years. The addition of semantic influence as features to a model once again improves the predictive performance of the model over baselines. In summary, our contributions are as follows:\footnote{The code and relevant data from our paper can be found at \url{http://github.com/sandeepsoni/contextual-leadership}}
\begin{itemize}
    \item We empirically demonstrate a link between long-term citation count and short-term linguistic influence, using both regression analysis (\autoref{sec:results-regression}) and an online prediction task (\autoref{sec:results-prediction}). 
    \item We present a method to estimate semantic influence using a parametric Hawkes process (\autoref{sec:methods-hawkes}). To achieve this, we find semantic changes and convert the usage of each change into a cascade (\autoref{sec:methods-cascades}). We also show that the method can be applied to quantify lexical influence.
    \item We present a method to identify monotonic semantic changes from timestamped text using contextual embeddings (see \autoref{sec:methods-contextual}).
\end{itemize}

\section{Methodology}
\label{sec:methodology}
This section describes our method for estimating the linguistic influence of each document in a timestamped collection. Our work builds on the theory of point process models~\citep{daley2003introduction}, in which the basic unit of data is a set of marked event timestamps. In our case, the events correspond to the use of an innovative word or usage; the mark corresponds to the document in which word or usage appears. To estimate linguistic influence of individual documents, we fit a parametric model in which per-document influence parameters explain the density of events in subsequent documents. We first describe the modeling framework in which these influence parameters are estimated (\autoref{sec:methods-hawkes}) and then describe how event cascades are constructed (\autoref{sec:methods-cascades}) from semantic changes (\autoref{sec:methods-contextual}) and lexical innovations (\autoref{sec:methods-lexical}).

\subsection{Estimating document influence from timestamped events}
\label{sec:methods-hawkes}
A marked cascade is a set of marked events $\{e_1, e_2, \ldots, e_{N}\}$, in which each event $e_i = (t_i, p_i)$ corresponds to a tuple of a timestamp $t_i$ and a mark $p_i$. Assume a set of marked cascades, indexed by $w \in \mathcal{W}$, with each mark belonging to a finite set that is shared across all cascades. In our application, each cascade corresponds to the appearances of an individual word or word sense, and each mark is the identity of the document in which the word or word sense appears. 

Point process models define probability distributions over cascades. In an inhomogeneous point process, the distribution of the count of events between any two timestamps $(t_1, t_2)$ is governed by the integral of an intensity function $\lambda(t, w)$. A Hawkes process is a special case in which the intensity function is the sum of terms associated with previous events~\citep{hawkes1971spectra}. We choose the following special form,
\begin{equation}
    \lambda(t, w) = c_w + \sum_{i: t_i^{(w)} < t} \alpha_{p_i^{(w)}} \kappa(t - t_i^{(w)}),
\label{eq:hp-intensity}
\end{equation}
where $\kappa$ is a time-decay kernel such as the exponential kernel $\kappa(\Delta t) = e^{-\gamma \Delta t}$ and $c_w$ is a constant. The parameter of interest is $\alpha$, which quantifies the influence exerted by the document $p_i^{(w)}$ on subsequent events.\footnote{In the more general multivariate Hawkes process, the intensity function can depend on the identity of ``receiver'' of influence. This enables the estimation of pairwise excitation parameters $\alpha_{i,j}$, as in the work of \citet{lukasik-etal-2016-hawkes}, to give an example from the NLP literature. However, it would be difficult to estimate pairwise excitation between thousands of documents, as required by our setting.}

Our application focuses on research papers, which historically have been published in a few bursts --- at conferences and in journals --- rather than continuously over time. For this reason we simplify our setting further, discretizing the timestamps by year. The evidence to be explained is now of the form $n(t, w)$, the count of word or sense $w$ in year $t$. We model this count as a Poisson random variable, and estimate the parameters $c_w$ and $\alpha$ by maximum likelihood. 

\subsection{Building event cascades}
\label{sec:methods-cascades}
To estimate the parameters in \autoref{eq:hp-intensity}, we require a set of timestamped events. Ideally these events should correspond to evidence of linguistic innovation. We consider two sources of events: semantic innovations (here focusing on words whose meaning changes over time) and lexical innovations (words whose usage rate increases dramatically over time).

We now introduce some notation used in the remainder of this section. Let a document be a sequence of discrete tokens from a finite vocabulary $\Vcal$, so that document $i$ is denoted $X_{i} = [x_{i}^{(1)}, x_{i}^{(2)}, \dots , x_{i}^{(n_i)}]$, with $n_i$ indicating the length of document $i$. A corpus is similarly defined as a set of $N$ documents, $\Xcal = \{X_1, X_2, . . . , X_N \}$, with each document associated with a discrete time $t_i \in \Tcal$.


\subsubsection{Using contextual embeddings to identify semantic changes}
\label{sec:methods-contextual}
We use contextual embeddings to identify words whose meaning changes over time, following prior work on computational historical linguistics~\citep[e.g.,][see \autoref{sec:related} for a more comprehensive review]{kutuzov-giulianelli-2020-uio}. A contextual embedding $\hbf_i^{(k)} \in \mathbb{R}^D$ is a vector representation of token $k$ in document $i$, computed from a model such as BERT~\citep{devlin-etal-2019-bert}. When the distribution over $\hbf$ for a given word changes over time, this is taken as evidence for a change in the word's meaning.

Let $\avg{t^{-}}{w}$ and $\avg{t^{+}}{w}$ be the count of the word $w$ up to and after time $t$, respectively. Specifically,
\begin{align*}
    \avg{t^{-}}{w} = \sum\limits_{i:t_i \le t} \sum\limits_{k}^{n_i} \mathds{1}(x_{i}^{(k)}=w)\\
    \avg{t^{+}}{w} = \sum\limits_{i:t_i > t} \sum\limits_{k}^{n_i} \mathds{1}(x_{i}^{(k)}=w)\\
\end{align*}
Average representations of the word $w$ up to and after time $t$, respectively, are calculated as follows.
\begin{align*}
    \avgrep{\vbf}{t^{-}}{w} = \frac{1}{\avg{t^{-}}{w}}\sum\limits_{i:t_i \le t} \sum\limits_{k}^{n_i} \hbf_i^{(k)} \mathds{1}(x_{i}^{(k)}=w)\\
    \avgrep{\vbf}{t^{+}}{w} = \frac{1}{\avg{t^{+}}{w}}\sum\limits_{i:t_i > t} \sum\limits_{k}^{n_i} \hbf_i^{(k)} \mathds{1}(x_{i}^{(k)}=w)\\
\end{align*}

Further, the variance in the contextual embeddings of the word $w$ over the entire corpus is calculated by taking the variance of each component of the embedding,
\begin{equation}
    \mathbf{s}_w = \frac{1}{m_w} \sum_{i, k: x_i^{(k)} = w}
    \left(\hbf_i^{(k)} - \boldsymbol{\mu}_w \right)^2,
\end{equation}
with $\mu_w$ equal to the mean contextualized embedding of word $w$.

A semantic change score for a word $w$ for a time $t$ is then the variance-weighted squared norm of the difference between its average pre-$t$ and post-$t$ contextualized embeddings (also known as the squared Mahalanobis distance):
\begin{equation}
    r(w,t) = (\avgrep{\vbf}{t^{-}}{w} - \avgrep{\vbf}{t^{+}}{w})^{\top} \Sbf_w^{-1} (\avgrep{\vbf}{t^{-}}{w} - \avgrep{\vbf}{t^{+}}{w}),
\label{eq:mahalanobis-metric}
\end{equation}
with $\Sbf_w = \text{Diag}(\mathbf{s}_{w}).$

\paragraph{Correction for frequency effects.} 
Both the mean and variance are estimated with larger samples for timestamps in the middle of $\Tcal$ in comparison to the initial and final timestamps. Consequently, the distance metric suffers from high sample variance for values of $t$ near these endpoints. The discrepancy is corrected by replacing the diagonal covariance $\Sbf_w$ in \autoref{eq:mahalanobis-metric} with an alternative covariance $\tilde{\Sbf}_w$ that reflects that additional uncertainty due to sample size. Specifically, we approximate the standard error of the mean $v_{t^-}$ as $\sqrt{\Sbf /m_{t^-}}$, and analogously for $v_{t^+}$. Then $\tilde{\Sbf}_w$ is defined as the product of these two approximate standard errors,
\begin{equation}
\label{eq:corrected-mahalanobis-metric}
    \tilde{\Sbf}_w = 
    \sqrt{\frac{\Sbf_w}{m_{t^{-},w}}} \sqrt{\frac{\Sbf_w} {m_{t^{+},w}}} 
    = \frac{\Sbf_w}{\sqrt{m_{t^{-},w} m_{t^{+},w}}}.
\end{equation}

Finally, $t^{*} = \argmax_{t} r(w,t)$ is selected as the transition point for the change in meaning of $w$. The changes are identified by sorting the words by $\max r(w, t)$ and applying a set of basic filters explained in \autoref{sec:experiments}. 
To give some intuition:
\begin{itemize}
    \item If $w$ changes in meaning at time $t$, then the difference in its representation up to $t$ and after $t$ should be high. The metric in \autoref{eq:mahalanobis-metric} captures this precisely by calculating the term $\avgrep{\vbf}{t^{-}}{w} - \avgrep{\vbf}{t^{+}}{w}$.
    \item Difference in average embeddings can be high for seasonal or bursty changes seen in words such as \example{turkey} which is referred to the bird more frequently at the time of American holidays~\citep{shoemark-etal-2019-room}. Rescaling the difference by the inverse variance encourages detection of monotonic changes.
    \item For rare words, the mean embeddings will be less reliable. The $\sqrt{m}$ terms in $\tilde{\Sbf}$ have the effect of emphasizing high-frequency words for which changes in the mean embedding are likely to be significant.
\end{itemize}

\paragraph{Distinguishing old and new usages.}
The previous step yields semantic innovations and their transition time. Simply identifying semantic changes is insufficient, since at any given time a word could be used in its old or new sense with respect to its time of transition. 
To categorize every usage of a semantic innovation $w$, the contextual embeddings are passed through a logistic regression classifier that predicts whether the usage is before or after the transition time. At the end of this step a sequence of embeddings for any semantic innovation is converted to a sequence of binary labels denoting their usage. For each word $w$, the cascade $(e_1^{(w)}, e_2^{(w)}, \ldots e_{N_w}^{(w)})$ is formed by filtering the usages to those that are classified as corresponding to the newer sense, with each event $e_i^{(w)}$ containing a timestamp $t^{(w)}_i$ and a document identifier $p^{(w)}_i$. These cascades are the evidence from which we estimate the per-document \emph{semantic} influence scores $\alpha_s$, as described in \autoref{sec:methods-hawkes}.

\paragraph{Why contextual embeddings?} Embeddings provide a powerful tool for understanding language change, offering more linguistic granularity than measures of change in the strength or composition of latent topics~\citep[e.g.,][]{griffiths2004finding,gerow2018measuring}. Prior work has employed diachronic non-contextual embeddings~\citep[e.g.,][]{soni2021follow}. Such methods require each word to have a single shared embedding in each time period. During periods in which a word is used in multiple senses, the non-contextual embedding must average across these senses, making it harder to detect changes in progress.

\subsubsection{Identifying lexical changes}
\label{sec:methods-lexical}
Unlike semantic changes, whose identification requires representations such as contextual embeddings, lexical changes are identified simply by comparing frequency changes. Specifically, for every word in a vocabulary we vary the segmentation year, say $t$, for the word and calculate the relative frequency up to and after $t$. We then take the best relative frequency ratio across the years as the score of lexical change for that word and aggregate to form a list of changes by sorting on this score. In contrast to semantic changes, all the usages of lexical changes are used to form cascades. These cascades are the evidence from which we estimate the per-document \emph{lexical} influence scores $\alpha_{\ell}$, again using the methods in \autoref{sec:methods-hawkes}.

\subsection{Overview}
To summarize the method for computing semantic influence:
\begin{enumerate}
    \item Compute the score $r(w,t)$ for each word $w$ and time $t$ as described in \autoref{eq:mahalanobis-metric} (with the adjusted covariance term from \autoref{eq:corrected-mahalanobis-metric}), and threshold to identify semantic changes.
    \item For each word selected in the previous step, classify each usage as either ``old'' or ``new'', and build a cascade from the timestamps of the new usages.
    \item Aggregating over all the cascades, estimate the influence parameters $\alpha_i$ for each document in the collection.
\end{enumerate}

\begin{figure*}
    \centering
    \includegraphics[width=.75\linewidth]{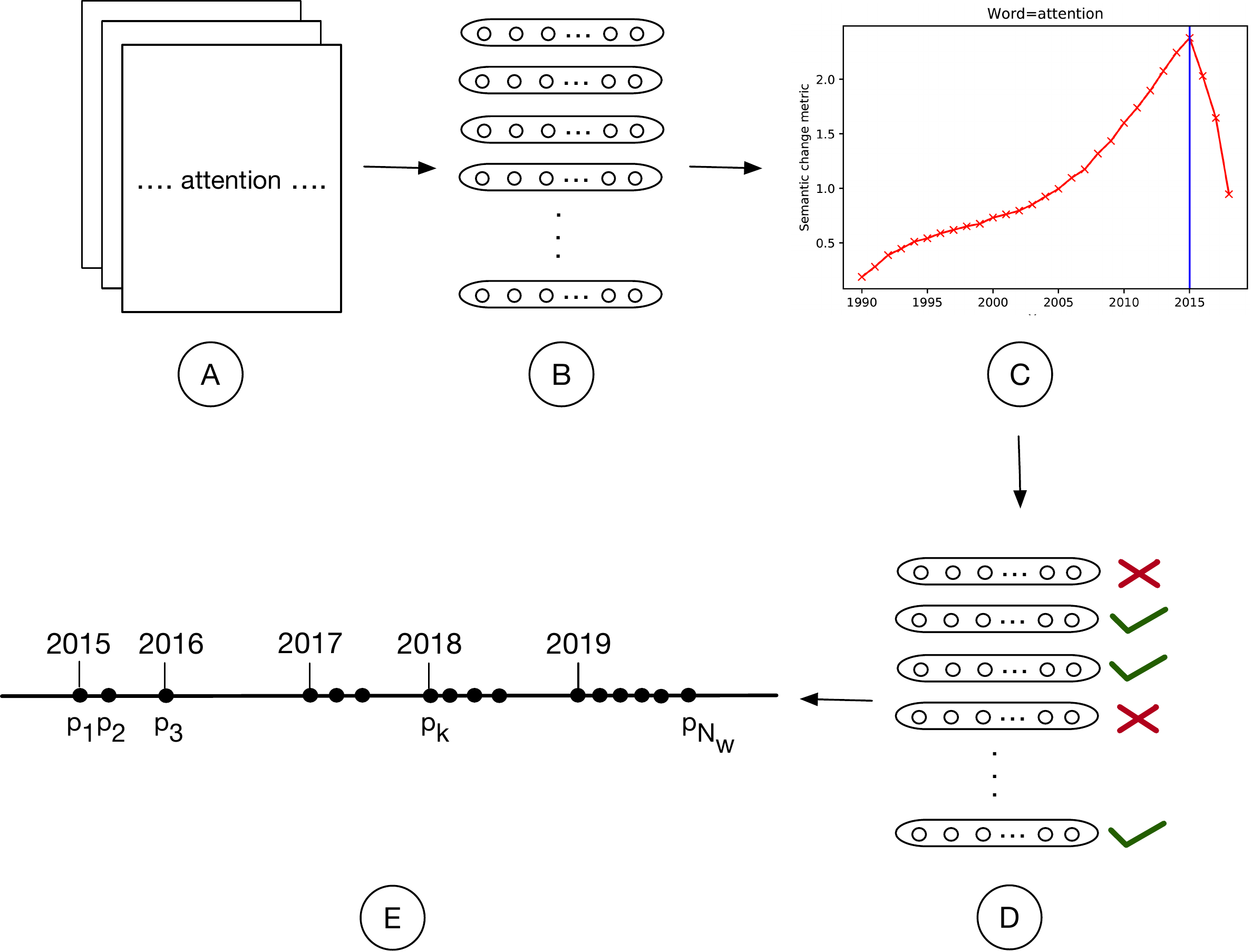}
    \caption{\textbf{Methodological pipeline}. The steps in our method can be summarized as follows for an example word \example{attention}. \textbf{(A)} depicts a collection of research papers that mention \example{attention}; \textbf{(B)} is a collection of contextual embeddings for \example{attention} across the entire corpus; \textbf{(C)} uses the contextual embeddings to find the transition point and the magnitude of the change; \textbf{(D)} uses the contextual embeddings to classify usages as old (marked with red crosses) or new (marked with green ticks) with respect to the transition time; \textbf{(E)} is a depiction of the event cascades comprising of timestamp and paper\_id ($p_i$) pairs.
    }
    \label{fig:pipeline}
\end{figure*}

A visual summary of the entire methodological pipeline is given in~\autoref{fig:pipeline}.

\section{Data}
\label{sec:data}

To construct a collection of research papers, we focus on papers that are included in the ACL anthology. We collected the ACL anthology bibliography file\footnote{Downloaded from \url{https://aclanthology.org/} on 12/9/2021} and converted the bib entries from the file as \texttt{JSON} objects; we retained the title of the paper, the year in which it was published, and the venue.

We then stripped all whitespace and special characters from the title of the paper. These stripped titles and the year of publication are matched with papers in \textit{s2orc} corpus~\citep{lo-etal-2020-s2orc}\footnote{20200705v1 version}. Matched papers that have a valid pdf parse (i.e full text of the paper) are retained. Though the s2orc dataset contains papers from as far back as $1965$, the coverage in the early years is sparse with few or no papers in many of the early years. As a result, the data is further filtered to retain only the papers that appear from $1990$ to $2019$ ($\Tcal = [1990, 2019]$). Descriptive statistics of the curated corpus is given in~\autoref{tab:descriptive-s2orc}.



\begin{table}
    \centering
    \begin{tabular}{@{}p{0.65\linewidth}p{0.3\linewidth}@{}}
    \toprule
    Number of papers & $36645$ \\
    Years  & $1990$--$2019$ \\
    Mean (median) cites per paper  & $6.68$ $(1)$ \\
    Mean (median) words per paper & $3291$ $(3248)$ \\
    \bottomrule
    \end{tabular}
    
    \caption{\textbf{Dataset summary.} Descriptive summary of the curated ACL corpus from \textit{s2orc} dataset.}
    \label{tab:descriptive-s2orc}
\end{table}
\section{Experimental Setup}
\label{sec:experiments}
For this study, multilingual \bert{} is used as the contextualizing model even though our data is English papers. This is to handle even those English language papers that have foreign language tokens. Specifically, the \texttt{bert-base-multilingual-uncased} model from the Hugging face~\citep{wolf-etal-2020-transformers} library is used.\footnote{\url{https://huggingface.co/bert-base-multilingual-uncased}} The size of the contextualized embeddings is $3144$ dimensions after concatenating the final four layers.

\paragraph{Continued pretraining} Previous work has shown that the quality of the contextual embeddings improves when the pretrained \bert{} is further trained on domain-specific text~\citep[e.g.,][]{gururangan-etal-2020-dont}. For this study, we continued to pretrain \bert{} model for $3$ epochs to optimize the masked language modeling objective. The probability of masking is set to 15 \%.

\paragraph{Wordpiece aggregation} Since \bert{} learns subword embeddings by breaking tokens into wordpieces, the embeddings of the wordpieces need to be aggregated to get a representation of a token. This aggregation is done by taking the average of the wordpiece embeddings.\footnote{Elementwise $\max$ as an alternative strategy of aggregation was also tried and performed similar in detecting changes.}

\paragraph{Data preprocessing} Non-English papers in the corpus are ignored from the analysis by identifying the language of the papers using \texttt{langid}~\citep[][]{lui-baldwin-2012-langid}. The vocabulary $\Vcal$ is constructed by retaining words that appear at least $10$ times in the abstracts and do not appear in more than 90 \% abstracts. Each paper is first segmented by whitespace and then broken into chunks of $200$ tokens. Only alphabetic tokens are retained. 

\paragraph{Classifying individual usages of semantic innovations} The off-the-shelf logistic regression classifier from \texttt{scikit-learn} is used to mark every individual instance of a semantic innovation as new or old. To avoid overfitting, we use $l_2$ regularization; all other inputs to the classifier are set to default. 4-fold cross-validation is performed to get the final assignment of labels from the classifier.

\paragraph{Word filters} We keep words in our vocabulary if they are composed only of alphabetic characters, occur in at most 90\% of the papers, and occur a minimum of $30$ times in the entire corpus. We also eliminate words whose length is less than or equal to $2$ characters. 

\paragraph{Estimation} To estimate the parameters of the Hawkes process, we use \texttt{scipy.optimize}, which internally uses the \texttt{L-BFGS} solver.
\section{Results}
\label{sec:results}

\subsection{Semantic changes}
We identified $2910$ semantic changes that capture several technical concepts in language research. The top changes and the period in which their meanings shift are shown in Table~\ref{tab:semantic-changes-examples}.

The evolution of language research, from the earlier focus on syntax and sequence processing using latent variable models to the current paradigm of using deep learning, is neatly summarized by the semantic innovations that the method identifies. Changes such as \example{tokenization} and \example{transducers} from the late nineties are indicative of the then-structural approach to core NLP research. 

The earlier part of the 2000s saw changes in terms such as \example{plan} (see \autoref{tab:semantic-changes-s2orc} for context in which the term appears), whose narrow usage in messaging applications broadened to other applications. The next decade also saw changes in terms such as \example{kernel} and \example{probabilistic}. These indicate the methodological changes that were underway during this period, with NLP research being dominated by a mix of kernel and bayesian methods during this decade~\citep[e.g.,][]{moschitti-2004-study,blei2003latent}. Methodological innovations such as conditional random fields~\citep[][]{lafferty2001conditional} and the rise of domain adaptation~\citep[e.g.,][]{chelba-acero-2004-adaptation,daume-iii-2007-frustratingly} are also evidenced by terms such as \example{conditional} and \example{adaptation}.

With the rise of neural approaches, words such as \example{representations}, \example{network}, and \example{decoder} underwent semantic changes between the years $2013$ to $2017$. Another prominent example of this shift is the term \example{attention}, shown in \autoref{fig:attention-example}, which shifts from its standard, broad usage to the more technical and focused usage with respect to neural networks around $2015$.

\begin{table}[]
    \centering
    \begin{tabular}{@{}p{0.22\linewidth} p{0.7\linewidth}@{}}
    \toprule

    Period & Semantic Changes\\

    \midrule

    2000-2002 & \example{system}, \example{data}, \example{plan}, \example{language}, \example{sentence}\\[4pt]

    2003-2005 & \example{state}, \example{task}, \example{relation}, \example{development}, \example{shared}\\[4pt]

    2006-2008 & \example{event}, \example{topic}, \example{comments}, \example{points}, \example{side}\\[4pt]

    2009-2011 & \example{media}, \example{user}, \example{social}, \example{users}, \example{neural}\\[4pt]

    2012-2014 & \example{network}, \example{hidden}, \example{embedding}, \example{layer}, \example{representations}\\[4pt]

    2015-2017 & \example{attention}, \example{representation}, \example{sequence}, \example{mechanism}, \example{decoder}\\[4pt]

    2018-2020 & \example{self}, \example{heads}, \example{glue}, \example{contextualized}, \example{attacks}\\[4pt]

    \bottomrule

    \end{tabular}
    \caption{\textbf{Examples of semantic changes.} We show a few handpicked examples amongst the top semantic changes in different periods. More context is shown in Table ~\ref{tab:semantic-changes-s2orc}.}
    \label{tab:semantic-changes-examples}
\end{table}

\subsection{Lexical changes}
We selected the top $3000$ lexical innovations to approximately match the number of semantic innovations. The lexical changes capture the introduction and rise in popularity of terms in language research. Unlike semantic changes, lexical changes are identified only by their change in frequency. 

Among the top changes are terms such as \example{bert}, \example{lstm}, \example{adam}, and \example{mturk} which are examples of new models, algorithms, tools, and technology introduced in language research. On the other hand, example changes such as \example{factuality}~\citep[e.g.,][]{sauri-pustejovsky-2012-sure,de-marneffe-etal-2012-happen,soni-etal-2014-modeling} and \example{sarcasm}~\citep{riloff-etal-2013-sarcasm,ptacek-etal-2014-sarcasm} indicate the rise in popularity of these concepts during specific years.

Abbreviations such as \example{sts} and \example{mt}, and names of languages such as \example{de} and \example{indonesian} are two categories of changes that prominently feature among top lexical changes. While the former indicates the necessity of naming technical concepts with memorable shortform, the latter is indicative of the rise in multilingual language research. 

\subsection{Regression analysis}
\label{sec:results-regression}
\begin{table*}

\centering
\small

\begingroup
\setlength{\tabcolsep}{6pt} 

\begin{tabular}{lp{.9cm}p{.9cm}p{.9cm}p{.9cm}}
\toprule

Predictors & M1 & M2 & M3 & M4\\

\midrule

Constant&-0.000&-0.080&-0.106&-0.116\\
 &(0.005)&(0.036)&(0.036)&(0.036)\\[4pt]
Initial Citations&0.763&0.740&0.727&0.718\\
 &(0.005)&(0.005)&(0.005)&(0.005)\\[4pt]
Lex. Inf. Q2& & &0.079&0.067\\
 & & &(0.012)&(0.012)\\[4pt]
Lex. Inf. Q3& & &0.086&0.064\\
 & & &(0.014)&(0.014)\\[4pt]
Lex. Inf. Q4& & &0.181&0.145\\
 & & &(0.017)&(0.018)\\[4pt]
Sem. Inf. Q2& & & &0.028\\
 & & & &(0.012)\\[4pt]
Sem. Inf. Q3& & & &0.091\\
 & & & &(0.015)\\[4pt]
Sem. Inf. Q4& & & &0.157\\
 & & & &(0.018)\\[4pt]
Log Lik.&-18828&-18681&-18615&-18569
\\
\bottomrule

\end{tabular}

\endgroup
\caption{\textbf{Regression analysis}. We show the results of long-term citations for various ablations. Each column indicates a model, each row indicates a predictor, and each cell contains the coefficient and, in parentheses, its standard error.  Topics are included as controls in models M2-4, but for clarity their coefficients are reserved for the supplementary material. Results for the best bandwidth parameter ($\gamma$=100), selected by the best heldout log-likelihood, are produced here whereas the regression results for other bandwidth settings are in the supplementary material.  
}
\label{tab:regress-citations}
\end{table*}
Our objective is to test whether the linguistic influence of a paper is positively correlated with its rate of future citations. However, many factors can confound our analysis including, but not limited to, the early citations a paper gets and the content of the paper. To control for these confounds and test our hypothesis, we frame the problem as a multivariate regression where features that proxy linguistic influence are incorporated alongside proxy features of other factors to predict the future citations. For our analysis in this section and~\autoref{sec:results-prediction}, we consider papers published in or after the year 2000, since the density of innovations appearing in these years is higher. The total number of papers in this interval is $19153$.

Our unit in the multivariate regression is a research paper and the dependent variable is the $Z$-normalized logarithm of its future citations. The $Z$-normalization uses a unique mean and variance for each year of publication, which helps to account for secular trends in the overall rate of citation over time. By ``future citations'' we mean the difference between the number of citations a paper gets five years after its publication (hereon referred as ``long-term citations'') and the number of citations a paper gets two years after its publication (hereon referred as ``short-term citations''). For example, for a paper published in $2012$, the short-term citations are from the period $2012-2014$ and the long-term citations are the citations accrued between $2015-2017$.

To test the impact of semantic influence, we include three baseline regression models. In the first baseline, M1, we include the $Z$-normalized short-term citations and a constant term as our only covariates. M2, our second baseline, consists of all covariates in M1 and the topic distribution of a paper learned from an LDA model~\citep{blei2003latent}. The topic distribution is taken as a coarse representation of the content of the paper.
Our final baseline is M3 which contains all the covariates from M2 in addition to categorical covariates corresponding to quantiles of the $Z$-normalized \emph{lexical} influence, $\alpha_l$, of each paper. We consider four quantiles: $< 50^{th}$ percentile, $\ge 50^{th}$ and $< 75^{th}$ percentile, $\ge 75^{th}$ percentile and $< 90^{th}$ percentile, and $\ge 90^{th}$ percentile. Finally, our experimental model, M4, has all the covariates from M3 and additional categorical covariates corresponding to the quantiles of the $Z$-normalized \emph{semantic} influence, $\alpha_s$, of each paper. The quantiles are divided in the same way as lexical influence.

The experimental model can be compared with the baseline models by their goodness-of-fit, measured by the log-likelihood of the data; analogously, the null hypothesis is that the goodness-of-fit of the experimental model is no better than that of the baseline models. Statistically, the likelihood ratio, our test statistic, follows a $\chi^2$ distribution with the excess number of parameters in the experimental model as the degrees of freedom. The null hypothesis can be rejected if the observed test statistic is determined to be unlikely under this distribution. 


The regression coefficients are shown in Table~\ref{tab:regress-citations}.\footnote{Due to space limitations we omit the topic coefficients from the table. The topics and their coefficients for M4 are shown in the appendix in \autoref{tab:topic-coefficients}.} Not surprisingly, short term citations are the strongest predictor of long-term citations, as seen by the strength of the regression coefficient. The regressions further reveal a strong relationship between semantic influence and  long-term citations: M4 obtains a significantly improved fit over M3, our strongest baseline ($\chi^2(3) = 91, p \approx 0.0$). 
Without additional controls, the average rate of long-term citations for the top quantile of semantic influence is $3$ times the long-term citation rate for the bottom quantile. With additional controls, the top quantile of semantic influence amounts to an increase in the expected citations by a factor of $1.2$, in comparison to the papers in the bottom quantile.

\subsection{Predicting future citations}
\label{sec:results-prediction}
We now turn to predicting the long-term citations from semantic influence and the other predictors described in~\autoref{sec:results-regression}. To more closely match the scenario of true future prediction, we formulate this as an online prediction task, in which the model is trained on past data to make predictions about future events~\citep{10.1162/COLI_a_00230,sogaard-etal-2021-need}. Formally, to make predictions about papers published in year $t$, we use information from the interval $[t, t+2]$ to compute the predictors: short-term citations, lexical influence, and semantic influence. We then make predictions about citations in years $[t+3, t+5]$. To estimate the weights of these predictors, we assume access to training data up to year $t+2$. We then increment $t$ and make predictions about the papers published in the next year. In this way, all papers published in the period 2001-2014 appear in one of the test folds.

The rest of the setup is similar to~\autoref{sec:results-regression} except one important difference. For the prediction task, we plug in estimates of lexical and semantic influence for all the values of $\gamma=\{0.001,0.01,0.1,1.0,10.0,100.0\}$ as predictors in the model. 
The results of the online prediction of long-term citations are shown in~\autoref{tab:predict-citations}. The performance is measured using mean squared error (MSE) between the predicted and ground-truth values. 
The model M4, which includes our measure of semantic influence, achieves the lowest error in 13 of 14 years, and it gives a more accurate prediction than M3 for 57.8\% of the 18554 papers in this slice of the dataset. 



\begin{table}

\centering
\small

\begingroup
\setlength{\tabcolsep}{6pt} 

\begin{tabular}{lrrrr}
\toprule
Publication Year & M1 & M2 & M3 & M4\\[2pt]

\midrule

2001 & 0.739 & 0.737 & 0.732 & 0.731\\[2pt]

2002 & 0.759 & 0.757 & 0.755 & 0.754\\[2pt]

2003 & 0.681 & 0.679 & 0.673 & 0.674\\[2pt]

2004 & 0.623 & 0.622 & 0.613 & 0.606\\[2pt]

2005 & 0.57 & 0.568 & 0.554 & 0.54\\[2pt]

2006 & 0.583 & 0.581 & 0.565 & 0.548\\[2pt]

2007 & 0.504 & 0.501 & 0.501 & 0.486\\[2pt]

2008 & 0.517 & 0.515 & 0.506 & 0.491\\[2pt]

2009 & 0.481 & 0.479 & 0.475 & 0.473\\[2pt]

2010 & 0.516 & 0.516 & 0.508 & 0.497\\[2pt]

2011 & 0.49 & 0.489 & 0.482 & 0.476\\[2pt]

2012 & 0.525 & 0.524 & 0.519 & 0.511\\[2pt]

2013 & 0.511 & 0.51 & 0.505 & 0.498\\[2pt]

2014 & 0.445 & 0.444 & 0.431 & 0.423\\[2pt]

\midrule

All Years & 0.529 & 0.528 & 0.52 & 0.511\\[2pt]

\\
\bottomrule

\end{tabular}

\endgroup
\caption{\textbf{Online predictive analysis} We show the performance in terms of MSE for the ablated models on the online citation prediction task. The first column indicates the publication year, the subsequent columns are the various ablations as seen in~\autoref{tab:regress-citations}, and each cell shows the MSE. The last row is the micro-averaged MSE over all the examples. Note that smaller values indicate better predictive performance.} 
\label{tab:predict-citations}
\end{table}

\section{Related Work}
\label{sec:related}

\subsection{Linguistic change and influence}
Several computational methods have been developed to identify changes in language~\citep{eisenstein-2019-measuring}. Of particular interest are techniques for detecting semantic changes in a text corpus. Such techniques are based on a range of representations, including frequency statistics~\citep[e.g.,][]{bybee2007diachronic}, static, type-level word embeddings~\citep[e.g.,][]{sagi-etal-2009-semantic,wijaya2011understanding,kulkarni2015statistically, hamilton-etal-2016-diachronic}, and contextual word embeddings~\citep[e.g.,][]{kutuzov-giulianelli-2020-uio,giulianelli-etal-2020-analysing, montariol-etal-2021-scalable}. Here, we use contextual embeddings which are, in principle, advantageous over static embeddings as they can distinguish the dynamics of co-existing senses.

Although there are many methods to detect changes, only a few computational studies find leaders or followers of these changes, which is important in order to understand who carries influence. By modeling lexical changes as cascades on a network, researchers have inferred that they propagate because of influence from strong ties~\citep[e.g.,][]{goel2016social}. Other researchers have identified leaders and followers of individual semantic changes and aggregated them to induce a leadership network between the sources~\citep{soni2021abolitionist}. Our work shares similarities with these prior studies but is distinct: We use similar cascade modeling techniques but for semantic changes, which are considerably harder to construct. 

Most relevant to our current work is that of \citet{soni2021follow} who find that semantically progressive scientific research papers get more citations. Semantic progressiveness --- a measure of linguistic novelty --- is calculated by comparing the old meaning of semantic innovations with their contemporary meaning in the context of the document. Our current work is different from this prior work in a key aspect: We estimate and establish a link between citation influence and semantic influence, instead of semantic novelty.


\subsection{Citation influence}
Citation count has historically been used as a proxy for the influence of a scientific article~\citep{fortunato2018science}, of researchers~\citep{borner2004simultaneous}, and is shown to be strongly correlated with scientific prestige~\citep{cole1968visibility}.Relevant to our work are studies that establish a link between citation influence to different measures of linguistic progressiveness. \citet{kelly2018measuring} find that progressiveness as measured in terms of difference in textual similarity between old and new patents is predictive of future citations of a patent. Similarly, \citet{soni2021follow} find that progressiveness measured as the early adoption of words of with newer meanings is predictive of citations of a paper. In contrast, in this paper, we find a link between linguistic influence in the short term to the future citations of the paper.
\section{Conclusion}
We have presented a new technique for quantifying semantic influence in time-stamped documents. Quantitative analysis demonstrates that this measure of semantic influence is strongly correlated with long-term citations a paper receives, and leads to improvement in the prediction of future citations. Our tool offers additional granularity in terms of linguistic influence, which can supplement structural measures of influence based on citation counts. 
Though we present quantitative analyses for scholarly documents in computational linguistics, our tool could be applied to scholarly documents in other research areas or to documents such as patents or court opinions where citation counts are considered structural measures of influence. We plan to focus on these applications in the future.

\section{Limitations}
\label{sec:limitations}
A simplifying assumption in this paper is there exists one dominant sense of a change before and after the transition point. This assumption may not hold for every change, in general, but helps in developing computational methods to identify a large array of changes. In future work, we plan to extend the ability of our method to identify co-evolving senses.

A fundamental limitation of the Hawkes Process is the closed-world assumption that all events are attributable to other observed events. This limitation is particularly relevant to our setting, where we observe only papers published in the ACL anthology, but those papers influence and are influenced by a much wider discourse, which includes not only other academic research papers but also software artifacts, books, and social media. In practice, this means that our method might wrongly assign credit to ``fast follower'' papers that are the first to adopt ideas published outside the ACL universe. Similarly, we make no attempt to measure the extent to which ACL anthology papers influence writing that is published elsewhere.

More generally, we cannot show whether the relationship between linguistic influence and citations is causal. The temporal asymmetry ensures that the future citations are not themselves causes of linguistic influence, but we cannot exclude the possibility that there is a common cause for both phenomena. For example, it seems likely that factors such as the overall quality of the research and the fame of the authors both contribute to the extent to which a paper drives the adoption of linguistic features in the short term, and to the number of citations it receives in the long term. Our regression analysis includes control variables for some potential common causes, such as topics, but it is not possible to control for all other potential confounders. Hence, our analysis should be considered \textit{correlational} and \textit{not causal}. Future work could focus on establishing and quantifying a causal link between linguistic influence and citations.
\section{Ethics Statement}
\label{sec:ethics}
This paper offers a new tool for understanding scientific communication. 
Because this tool quantifies the linguistic impact of research papers, there is the possibility that it could be used for consequential decisions such as hiring, promotion, and funding. 
This implies a ``leaderboard'' approach to scholarship that would overvalue the most fashionable mainstream research topics, while penalizing research that has a deep impact in a relatively small community.
Similar concerns have been raised about other measures of academic impact: Jorge Hirsch, the inventor of the $H$-index, noted that his metric could have ``severe unintended negative consequences,'' and urged evaluators to go beyond any single index to consider the broader context when considering an individual's scientific contributions~\citep{conroy2020s}. The same applies to semantic influence metric defined in this paper.
\section*{Acknowledgements}
We thank David Mimno and William Cohen who gave feedback on the initial draft of this paper. We also thank all the anonymous reviewers for their time and suggestions which led to improvements of the paper. The research reported in this article was supported by funding from the National Science Foundation (IIS-1813470 and IIS-1942591).
\bibliography{anthology,custom}

\begin{thebibliography}{50}
\expandafter\ifx\csname natexlab\endcsname\relax\def\natexlab#1{#1}\fi

\bibitem[{Bahdanau et~al.(2015)Bahdanau, Cho, and Bengio}]{bahdanau2015neural}
Dzmitry Bahdanau, Kyung~Hyun Cho, and Yoshua Bengio. 2015.
\newblock Neural machine translation by jointly learning to align and
  translate.
\newblock In \emph{3rd International Conference on Learning Representations,
  ICLR 2015}.

\bibitem[{Blei et~al.(2003)Blei, Ng, and Jordan}]{blei2003latent}
David~M Blei, Andrew~Y Ng, and Michael~I Jordan. 2003.
\newblock Latent dirichlet allocation.
\newblock \emph{Journal of Machine Learning Research}, 3(Jan):993--1022.

\bibitem[{B{\"o}rner et~al.(2004)B{\"o}rner, Maru, and
  Goldstone}]{borner2004simultaneous}
Katy B{\"o}rner, Jeegar~T Maru, and Robert~L Goldstone. 2004.
\newblock The simultaneous evolution of author and paper networks.
\newblock \emph{Proceedings of the National Academy of Sciences}, 101(suppl
  1):5266--5273.

\bibitem[{Bornmann and Daniel(2008)}]{bornmann2008citation}
Lutz Bornmann and Hans-Dieter Daniel. 2008.
\newblock What do citation counts measure? a review of studies on citing
  behavior.
\newblock \emph{Journal of documentation}.

\bibitem[{Bybee(2007)}]{bybee2007diachronic}
Joan~L Bybee. 2007.
\newblock Diachronic linguistics.
\newblock In \emph{The Oxford handbook of cognitive linguistics}.

\bibitem[{Chelba and Acero(2004)}]{chelba-acero-2004-adaptation}
Ciprian Chelba and Alex Acero. 2004.
\newblock \href {https://aclanthology.org/W04-3237} {Adaptation of maximum
  entropy capitalizer: Little data can help a lo}.
\newblock In \emph{Proceedings of the 2004 Conference on Empirical Methods in
  Natural Language Processing}, pages 285--292, Barcelona, Spain. Association
  for Computational Linguistics.

\bibitem[{Cole and Cole(1968)}]{cole1968visibility}
Stephen Cole and Jonathan~R Cole. 1968.
\newblock Visibility and the structural bases of awareness of scientific
  research.
\newblock \emph{American sociological review}, pages 397--413.

\bibitem[{Conroy(2020)}]{conroy2020s}
Gemma Conroy. 2020.
\newblock What’s wrong with the h-index, according to its inventor.
\newblock \emph{Nature Index}, 24.

\bibitem[{Cronin(2005)}]{cronin2005hand}
Blaise Cronin. 2005.
\newblock \emph{The hand of science: Academic writing and its rewards}.
\newblock Scarecrow press.

\bibitem[{Daley et~al.(2003)Daley, Vere-Jones et~al.}]{daley2003introduction}
Daryl~J Daley, David Vere-Jones, et~al. 2003.
\newblock \emph{An introduction to the theory of point processes: volume I:
  elementary theory and methods}.
\newblock Springer.

\bibitem[{Daum{\'e}~III(2007)}]{daume-iii-2007-frustratingly}
Hal Daum{\'e}~III. 2007.
\newblock \href {https://aclanthology.org/P07-1033} {Frustratingly easy domain
  adaptation}.
\newblock In \emph{Proceedings of the 45th Annual Meeting of the Association of
  Computational Linguistics}, pages 256--263, Prague, Czech Republic.
  Association for Computational Linguistics.

\bibitem[{de~Marneffe et~al.(2012)de~Marneffe, Manning, and
  Potts}]{de-marneffe-etal-2012-happen}
Marie-Catherine de~Marneffe, Christopher~D. Manning, and Christopher Potts.
  2012.
\newblock \href {https://doi.org/10.1162/COLI_a_00097} {Did it happen? the
  pragmatic complexity of veridicality assessment}.
\newblock \emph{Computational Linguistics}, 38(2):301--333.

\bibitem[{Devlin et~al.(2019)Devlin, Chang, Lee, and
  Toutanova}]{devlin-etal-2019-bert}
Jacob Devlin, Ming-Wei Chang, Kenton Lee, and Kristina Toutanova. 2019.
\newblock \href {https://doi.org/10.18653/v1/N19-1423} {{BERT}: Pre-training of
  deep bidirectional transformers for language understanding}.
\newblock In \emph{Proceedings of the 2019 Conference of the North {A}merican
  Chapter of the Association for Computational Linguistics: Human Language
  Technologies, Volume 1 (Long and Short Papers)}, pages 4171--4186,
  Minneapolis, Minnesota. Association for Computational Linguistics.

\bibitem[{Dubossarsky et~al.(2019)Dubossarsky, Hengchen, Tahmasebi, and
  Schlechtweg}]{dubossarsky-etal-2019-time}
Haim Dubossarsky, Simon Hengchen, Nina Tahmasebi, and Dominik Schlechtweg.
  2019.
\newblock \href {https://doi.org/10.18653/v1/P19-1044} {Time-out: Temporal
  referencing for robust modeling of lexical semantic change}.
\newblock In \emph{Proceedings of the 57th Annual Meeting of the Association
  for Computational Linguistics}, pages 457--470, Florence, Italy. Association
  for Computational Linguistics.

\bibitem[{Eisenstein(2019)}]{eisenstein-2019-measuring}
Jacob Eisenstein. 2019.
\newblock \href {https://doi.org/10.18653/v1/N19-5003} {Measuring and modeling
  language change}.
\newblock In \emph{Proceedings of the 2019 Conference of the North {A}merican
  Chapter of the Association for Computational Linguistics: Tutorials}, pages
  9--14, Minneapolis, Minnesota. Association for Computational Linguistics.

\bibitem[{Fortunato et~al.(2018)Fortunato, Bergstrom, B{\"o}rner, Evans,
  Helbing, Milojevi{\'c}, Petersen, Radicchi, Sinatra, Uzzi, Vespignani,
  Waltman, Wang, and Barab{\'a}si}]{fortunato2018science}
Santo Fortunato, Carl~T. Bergstrom, Katy B{\"o}rner, James~A. Evans, Dirk
  Helbing, Sta{\v s}a Milojevi{\'c}, Alexander~M. Petersen, Filippo Radicchi,
  Roberta Sinatra, Brian Uzzi, Alessandro Vespignani, Ludo Waltman, Dashun
  Wang, and Albert-L{\'a}szl{\'o} Barab{\'a}si. 2018.
\newblock Science of science.
\newblock \emph{Science}, 359(6379).

\bibitem[{Gerow et~al.(2018)Gerow, Hu, Boyd-Graber, Blei, and
  Evans}]{gerow2018measuring}
Aaron Gerow, Yuening Hu, Jordan Boyd-Graber, David~M Blei, and James~A Evans.
  2018.
\newblock Measuring discursive influence across scholarship.
\newblock \emph{Proceedings of the national academy of sciences},
  115(13):3308--3313.

\bibitem[{Giulianelli et~al.(2020)Giulianelli, Del~Tredici, and
  Fern{\'a}ndez}]{giulianelli-etal-2020-analysing}
Mario Giulianelli, Marco Del~Tredici, and Raquel Fern{\'a}ndez. 2020.
\newblock \href {https://doi.org/10.18653/v1/2020.acl-main.365} {Analysing
  lexical semantic change with contextualised word representations}.
\newblock In \emph{Proceedings of the 58th Annual Meeting of the Association
  for Computational Linguistics}, pages 3960--3973, Online. Association for
  Computational Linguistics.

\bibitem[{Goel et~al.(2016)Goel, Soni, Goyal, Paparrizos, Wallach, Diaz, and
  Eisenstein}]{goel2016social}
Rahul Goel, Sandeep Soni, Naman Goyal, John Paparrizos, Hanna Wallach, Fernando
  Diaz, and Jacob Eisenstein. 2016.
\newblock The social dynamics of language change in online networks.
\newblock In \emph{International conference on social informatics}, pages
  41--57. Springer.

\bibitem[{Griffiths and Steyvers(2004)}]{griffiths2004finding}
Thomas~L Griffiths and Mark Steyvers. 2004.
\newblock Finding scientific topics.
\newblock \emph{Proceedings of the National academy of Sciences},
  101(suppl\_1):5228--5235.

\bibitem[{Gururangan et~al.(2020)Gururangan, Marasovi{\'c}, Swayamdipta, Lo,
  Beltagy, Downey, and Smith}]{gururangan-etal-2020-dont}
Suchin Gururangan, Ana Marasovi{\'c}, Swabha Swayamdipta, Kyle Lo, Iz~Beltagy,
  Doug Downey, and Noah~A. Smith. 2020.
\newblock \href {https://doi.org/10.18653/v1/2020.acl-main.740} {Don{'}t stop
  pretraining: Adapt language models to domains and tasks}.
\newblock In \emph{Proceedings of the 58th Annual Meeting of the Association
  for Computational Linguistics}, pages 8342--8360, Online. Association for
  Computational Linguistics.

\bibitem[{Hamilton et~al.(2016)Hamilton, Leskovec, and
  Jurafsky}]{hamilton-etal-2016-diachronic}
William~L. Hamilton, Jure Leskovec, and Dan Jurafsky. 2016.
\newblock \href {https://doi.org/10.18653/v1/P16-1141} {Diachronic word
  embeddings reveal statistical laws of semantic change}.
\newblock In \emph{Proceedings of the 54th Annual Meeting of the Association
  for Computational Linguistics (Volume 1: Long Papers)}, pages 1489--1501,
  Berlin, Germany. Association for Computational Linguistics.

\bibitem[{Hawkes(1971)}]{hawkes1971spectra}
Alan~G Hawkes. 1971.
\newblock Spectra of some self-exciting and mutually exciting point processes.
\newblock \emph{Biometrika}, 58(1):83--90.

\bibitem[{Inhaber and Przednowek(1976)}]{inhaber1976quality}
H~Inhaber and KJSSoS Przednowek. 1976.
\newblock Quality of research and the nobel prizes.
\newblock \emph{Social Studies of Science}, 6(1):33--50.

\bibitem[{Karimi et~al.(2015)Karimi, Yin, and Baum}]{10.1162/COLI_a_00230}
Sarvnaz Karimi, Jie Yin, and Jiri Baum. 2015.
\newblock \href {https://doi.org/10.1162/COLI_a_00230} {{Evaluation Methods for
  Statistically Dependent Text}}.
\newblock \emph{Computational Linguistics}, 41(3):539--548.

\bibitem[{Kelly et~al.(2018)Kelly, Papanikolaou, Seru, and
  Taddy}]{kelly2018measuring}
Bryan~T Kelly, Dimitris Papanikolaou, Amit Seru, and Matt Taddy. 2018.
\newblock Measuring technological innovation over the long run.
\newblock \emph{NBER Working Paper}, (w25266).

\bibitem[{Kim et~al.(2014)Kim, Chiu, Hanaki, Hegde, and
  Petrov}]{kim-etal-2014-temporal}
Yoon Kim, Yi-I Chiu, Kentaro Hanaki, Darshan Hegde, and Slav Petrov. 2014.
\newblock \href {https://doi.org/10.3115/v1/W14-2517} {Temporal analysis of
  language through neural language models}.
\newblock In \emph{Proceedings of the {ACL} 2014 Workshop on Language
  Technologies and Computational Social Science}, pages 61--65, Baltimore, MD,
  USA. Association for Computational Linguistics.

\bibitem[{Kulkarni et~al.(2015)Kulkarni, Al-Rfou, Perozzi, and
  Skiena}]{kulkarni2015statistically}
Vivek Kulkarni, Rami Al-Rfou, Bryan Perozzi, and Steven Skiena. 2015.
\newblock Statistically significant detection of linguistic change.
\newblock In \emph{Proceedings of the 24th International Conference on World
  Wide Web}, pages 625--635. International World Wide Web Conferences Steering
  Committee.

\bibitem[{Kutuzov and Giulianelli(2020)}]{kutuzov-giulianelli-2020-uio}
Andrey Kutuzov and Mario Giulianelli. 2020.
\newblock \href {https://doi.org/10.18653/v1/2020.semeval-1.14}
  {{U}i{O}-{U}v{A} at {S}em{E}val-2020 task 1: Contextualised embeddings for
  lexical semantic change detection}.
\newblock In \emph{Proceedings of the Fourteenth Workshop on Semantic
  Evaluation}, pages 126--134, Barcelona (online). International Committee for
  Computational Linguistics.

\bibitem[{Lafferty et~al.(2001)Lafferty, McCallum, and
  Pereira}]{lafferty2001conditional}
John~D Lafferty, Andrew McCallum, and Fernando~CN Pereira. 2001.
\newblock Conditional random fields: Probabilistic models for segmenting and
  labeling sequence data.
\newblock In \emph{Proceedings of the Eighteenth International Conference on
  Machine Learning}, pages 282--289.

\bibitem[{Lawani(1986)}]{lawani1986some}
Stephen Lawani. 1986.
\newblock Some bibliometric correlates of quality in scientific research.
\newblock \emph{Scientometrics}, 9(1-2):13--25.

\bibitem[{Lo et~al.(2020)Lo, Wang, Neumann, Kinney, and
  Weld}]{lo-etal-2020-s2orc}
Kyle Lo, Lucy~Lu Wang, Mark Neumann, Rodney Kinney, and Daniel Weld. 2020.
\newblock \href {https://doi.org/10.18653/v1/2020.acl-main.447} {{S}2{ORC}: The
  semantic scholar open research corpus}.
\newblock In \emph{Proceedings of the 58th Annual Meeting of the Association
  for Computational Linguistics}, pages 4969--4983, Online. Association for
  Computational Linguistics.

\bibitem[{Lui and Baldwin(2012)}]{lui-baldwin-2012-langid}
Marco Lui and Timothy Baldwin. 2012.
\newblock \href {https://aclanthology.org/P12-3005} {langid.py: An
  off-the-shelf language identification tool}.
\newblock In \emph{Proceedings of the {ACL} 2012 System Demonstrations}, pages
  25--30, Jeju Island, Korea. Association for Computational Linguistics.

\bibitem[{Lukasik et~al.(2016)Lukasik, Srijith, Vu, Bontcheva, Zubiaga, and
  Cohn}]{lukasik-etal-2016-hawkes}
Michal Lukasik, P.~K. Srijith, Duy Vu, Kalina Bontcheva, Arkaitz Zubiaga, and
  Trevor Cohn. 2016.
\newblock \href {https://doi.org/10.18653/v1/P16-2064} {{H}awkes processes for
  continuous time sequence classification: an application to rumour stance
  classification in {T}witter}.
\newblock In \emph{Proceedings of the 54th Annual Meeting of the Association
  for Computational Linguistics (Volume 2: Short Papers)}, pages 393--398,
  Berlin, Germany. Association for Computational Linguistics.

\bibitem[{Montariol et~al.(2021)Montariol, Martinc, and
  Pivovarova}]{montariol-etal-2021-scalable}
Syrielle Montariol, Matej Martinc, and Lidia Pivovarova. 2021.
\newblock \href {https://doi.org/10.18653/v1/2021.naacl-main.369} {Scalable and
  interpretable semantic change detection}.
\newblock In \emph{Proceedings of the 2021 Conference of the North American
  Chapter of the Association for Computational Linguistics: Human Language
  Technologies}, pages 4642--4652, Online. Association for Computational
  Linguistics.

\bibitem[{Moschitti(2004)}]{moschitti-2004-study}
Alessandro Moschitti. 2004.
\newblock \href {https://doi.org/10.3115/1218955.1218998} {A study on
  convolution kernels for shallow statistic parsing}.
\newblock In \emph{Proceedings of the 42nd Annual Meeting of the Association
  for Computational Linguistics ({ACL}-04)}, pages 335--342, Barcelona, Spain.

\bibitem[{Pt{\'a}{\v{c}}ek et~al.(2014)Pt{\'a}{\v{c}}ek, Habernal, and
  Hong}]{ptacek-etal-2014-sarcasm}
Tom{\'a}{\v{s}} Pt{\'a}{\v{c}}ek, Ivan Habernal, and Jun Hong. 2014.
\newblock \href {https://aclanthology.org/C14-1022} {Sarcasm detection on
  {C}zech and {E}nglish {T}witter}.
\newblock In \emph{Proceedings of {COLING} 2014, the 25th International
  Conference on Computational Linguistics: Technical Papers}, pages 213--223,
  Dublin, Ireland. Dublin City University and Association for Computational
  Linguistics.

\bibitem[{Riloff et~al.(2013)Riloff, Qadir, Surve, De~Silva, Gilbert, and
  Huang}]{riloff-etal-2013-sarcasm}
Ellen Riloff, Ashequl Qadir, Prafulla Surve, Lalindra De~Silva, Nathan Gilbert,
  and Ruihong Huang. 2013.
\newblock \href {https://aclanthology.org/D13-1066} {Sarcasm as contrast
  between a positive sentiment and negative situation}.
\newblock In \emph{Proceedings of the 2013 Conference on Empirical Methods in
  Natural Language Processing}, pages 704--714, Seattle, Washington, USA.
  Association for Computational Linguistics.

\bibitem[{Rosenfeld and Erk(2018)}]{rosenfeld-erk-2018-deep}
Alex Rosenfeld and Katrin Erk. 2018.
\newblock \href {https://doi.org/10.18653/v1/N18-1044} {Deep neural models of
  semantic shift}.
\newblock In \emph{Proceedings of the 2018 Conference of the North {A}merican
  Chapter of the Association for Computational Linguistics: Human Language
  Technologies, Volume 1 (Long Papers)}, pages 474--484, New Orleans,
  Louisiana. Association for Computational Linguistics.

\bibitem[{Sagi et~al.(2009)Sagi, Kaufmann, and Clark}]{sagi-etal-2009-semantic}
Eyal Sagi, Stefan Kaufmann, and Brady Clark. 2009.
\newblock \href {https://aclanthology.org/W09-0214} {Semantic density analysis:
  Comparing word meaning across time and phonetic space}.
\newblock In \emph{Proceedings of the Workshop on Geometrical Models of Natural
  Language Semantics}, pages 104--111, Athens, Greece. Association for
  Computational Linguistics.

\bibitem[{Saur{\'\i} and Pustejovsky(2012)}]{sauri-pustejovsky-2012-sure}
Roser Saur{\'\i} and James Pustejovsky. 2012.
\newblock \href {https://doi.org/10.1162/COLI_a_00096} {Are you sure that this
  happened? assessing the factuality degree of events in text}.
\newblock \emph{Computational Linguistics}, 38(2):261--299.

\bibitem[{Shoemark et~al.(2019)Shoemark, Liza, Nguyen, Hale, and
  McGillivray}]{shoemark-etal-2019-room}
Philippa Shoemark, Farhana~Ferdousi Liza, Dong Nguyen, Scott Hale, and Barbara
  McGillivray. 2019.
\newblock \href {https://doi.org/10.18653/v1/D19-1007} {Room to {G}lo: A
  systematic comparison of semantic change detection approaches with word
  embeddings}.
\newblock In \emph{Proceedings of the 2019 Conference on Empirical Methods in
  Natural Language Processing and the 9th International Joint Conference on
  Natural Language Processing (EMNLP-IJCNLP)}, pages 66--76, Hong Kong, China.
  Association for Computational Linguistics.

\bibitem[{S{\o}gaard et~al.(2021)S{\o}gaard, Ebert, Bastings, and
  Filippova}]{sogaard-etal-2021-need}
Anders S{\o}gaard, Sebastian Ebert, Jasmijn Bastings, and Katja Filippova.
  2021.
\newblock \href {https://doi.org/10.18653/v1/2021.eacl-main.156} {We need to
  talk about random splits}.
\newblock In \emph{Proceedings of the 16th Conference of the European Chapter
  of the Association for Computational Linguistics: Main Volume}, pages
  1823--1832, Online. Association for Computational Linguistics.

\bibitem[{Soni et~al.(2021{\natexlab{a}})Soni, Klein, and
  Eisenstein}]{soni2021abolitionist}
Sandeep Soni, Lauren~F Klein, and Jacob Eisenstein. 2021{\natexlab{a}}.
\newblock Abolitionist networks: Modeling language change in nineteenth-century
  activist newspapers.
\newblock \emph{Journal of Cultural Analytics}, 6(1):18841.

\bibitem[{Soni et~al.(2021{\natexlab{b}})Soni, Lerman, and
  Eisenstein}]{soni2021follow}
Sandeep Soni, Kristina Lerman, and Jacob Eisenstein. 2021{\natexlab{b}}.
\newblock Follow the leader: Documents on the leading edge of semantic change
  get more citations.
\newblock \emph{Journal of the Association for Information Science and
  Technology}, 72(4):478--492.

\bibitem[{Soni et~al.(2014)Soni, Mitra, Gilbert, and
  Eisenstein}]{soni-etal-2014-modeling}
Sandeep Soni, Tanushree Mitra, Eric Gilbert, and Jacob Eisenstein. 2014.
\newblock \href {https://doi.org/10.3115/v1/P14-2068} {Modeling factuality
  judgments in social media text}.
\newblock In \emph{Proceedings of the 52nd Annual Meeting of the Association
  for Computational Linguistics (Volume 2: Short Papers)}, pages 415--420,
  Baltimore, Maryland. Association for Computational Linguistics.

\bibitem[{Traugott and Dasher(2001)}]{traugott2001regularity}
Elizabeth~Closs Traugott and Richard~B Dasher. 2001.
\newblock \emph{Regularity in semantic change}, volume~97.
\newblock Cambridge University Press.

\bibitem[{Vaswani et~al.(2017)Vaswani, Shazeer, Parmar, Uszkoreit, Jones,
  Gomez, Kaiser, and Polosukhin}]{vaswani2017attention}
Ashish Vaswani, Noam Shazeer, Niki Parmar, Jakob Uszkoreit, Llion Jones,
  Aidan~N Gomez, {\L}ukasz Kaiser, and Illia Polosukhin. 2017.
\newblock Attention is all you need.
\newblock \emph{Advances in neural information processing systems}, 30.

\bibitem[{Wijaya and Yeniterzi(2011)}]{wijaya2011understanding}
Derry~Tanti Wijaya and Reyyan Yeniterzi. 2011.
\newblock Understanding semantic change of words over centuries.
\newblock In \emph{Proceedings of the 2011 international workshop on DETecting
  and Exploiting Cultural diversiTy on the social web}, pages 35--40.

\bibitem[{Wolf et~al.(2020)Wolf, Debut, Sanh, Chaumond, Delangue, Moi, Cistac,
  Rault, Louf, Funtowicz, Davison, Shleifer, von Platen, Ma, Jernite, Plu, Xu,
  Le~Scao, Gugger, Drame, Lhoest, and Rush}]{wolf-etal-2020-transformers}
Thomas Wolf, Lysandre Debut, Victor Sanh, Julien Chaumond, Clement Delangue,
  Anthony Moi, Pierric Cistac, Tim Rault, Remi Louf, Morgan Funtowicz, Joe
  Davison, Sam Shleifer, Patrick von Platen, Clara Ma, Yacine Jernite, Julien
  Plu, Canwen Xu, Teven Le~Scao, Sylvain Gugger, Mariama Drame, Quentin Lhoest,
  and Alexander Rush. 2020.
\newblock \href {https://doi.org/10.18653/v1/2020.emnlp-demos.6} {Transformers:
  State-of-the-art natural language processing}.
\newblock In \emph{Proceedings of the 2020 Conference on Empirical Methods in
  Natural Language Processing: System Demonstrations}, pages 38--45, Online.
  Association for Computational Linguistics.

\end{thebibliography}
\bibliographystyle{acl_natbib}

\appendix
\section{Examples of Semantic Changes}
\label{sec:semantic-change}
\begin{figure}
    \centering
    \includegraphics[width=\linewidth]{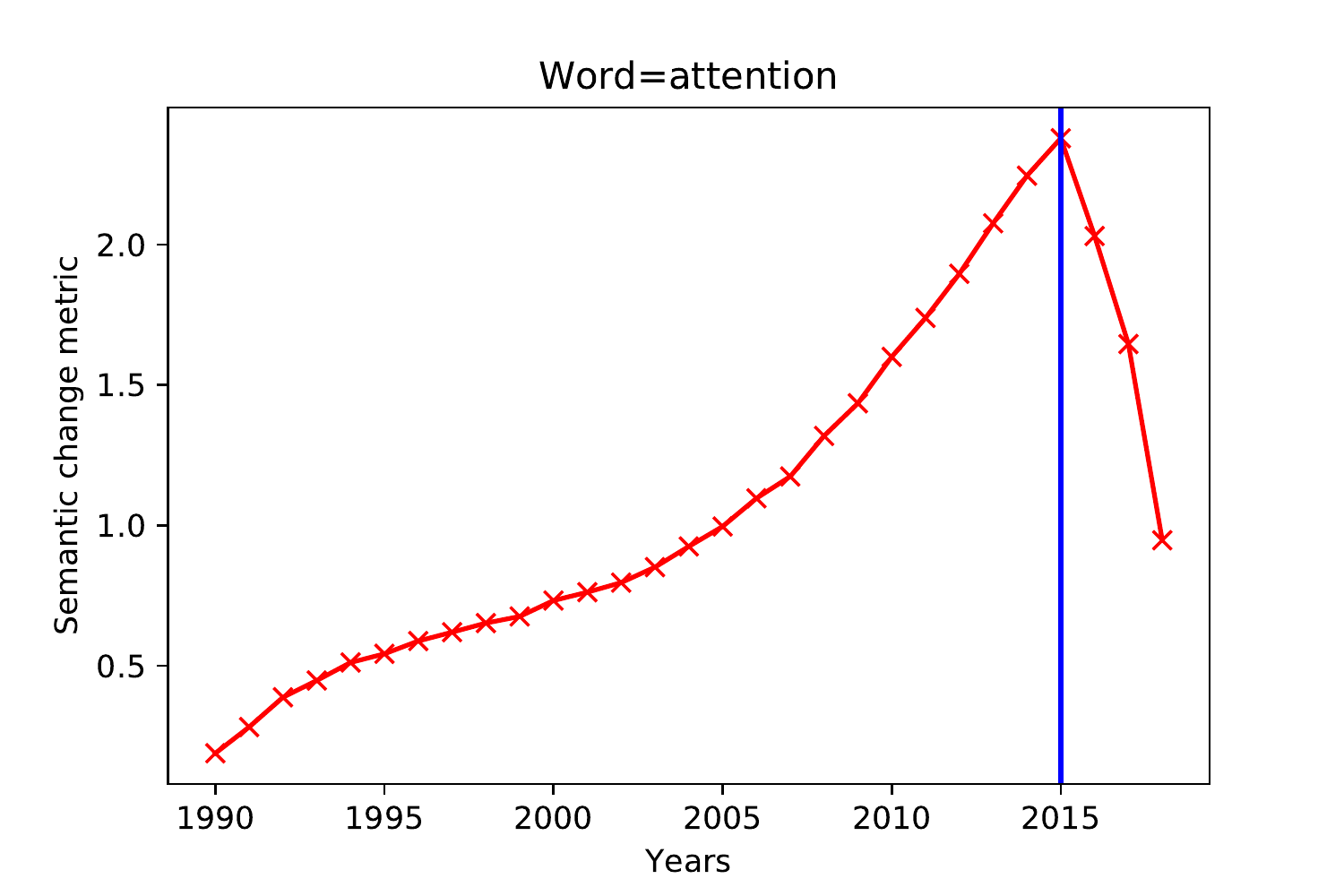}
    \caption{\textbf{Visual depiction of change in top example.} Semantic change in the term \example{attention} in \textit{s2orc}'s ACL anthology subset. The blue line indicates the transition year for meaning change. The transition year for the term \example{attention} coincides with early papers that described the attention mechanism in neural networks~\citep[][]{bahdanau2015neural} that later became the bedrock of transformers architecture~\citep[][]{vaswani2017attention}
    }
    \label{fig:attention-example}
\end{figure}
We show more statistical details for some of the semantic changes and the context in which these changes occur in~\autoref{tab:semantic-changes-s2orc}.
\begin{table*}
    \centering
    \begin{tabular}{@{}lllp{.6in}p{.6in}p{1.4in}p{1.4in}@{}}
    \toprule
    Term & Year & Score & Relative count pre-transition & Relative count post-transition & Earlier usages & Later usages\\
    \midrule
    \example{attention} & $2015$ & $2.38$ & $126$ & $1670$ & \example{increased \textbf{attention} over
the past several years} & \example{parallelizable \textbf{attention} networks}\\[1ex] 
    & & & & & \example{need to be paid \textbf{attention}} & \example{vector of \textbf{attention} weights}\\[2ex] 
    \example{plan} & $2001$ & $1.52$ & $381$ & $158$ & \example{\textbf{plan} such a message} & \example{\textbf{plan} recognition problems}\\[1ex]
    & & & & & \example{embedded in the \textbf{plan} library} & \example{\textbf{plan} for tag generation}\\[2ex]
    \example{network} & $2013$ & $1.19$ & $240$ & $1000$ & \example{semantic \textbf{network} path schemata} & \example{deep learning \textbf{network} configurations}\\[1ex]
    & & & & & \example{\textbf{network} of semantically related noun senses} & \example{\textbf{network} parameters to tune}\\[2ex]
    \example{focus} & $2006$ & $0.99$ & $451$ & $521$ & \example{tracking local \textbf{focus}} & \example{main \textbf{focus} of our work}\\[1ex]
    & & & & & \example{\textbf{focus} of attention in discourse} & \example{the \textbf{focus} particle}\\[2ex]
    \example{representations} & $2013$ & $0.94$ & $257$ & $1018$ & \example{grammatical \textbf{representations}} & \example{learning distributed \textbf{representations}}\\[1ex]
    & & & & & \example{logical semantic \textbf{representations}} & \example{learned \textbf{representations} across views}\\[2ex]
    \example{deep} & $2014$ & $0.94$ & $114$ & $417$ & \example{\textbf{deep} cognitive understanding} & \example{\textbf{deep} learning}\\[1ex]
    & & & & & \example{\textbf{deep} syntactic features} & \example{\textbf{deep} architectures}\\[2ex]    
    \bottomrule
    \end{tabular}
    \caption{\textbf{Semantic change examples.} Top examples of semantic changes identified from the curated ACL corpus from the \textit{s2orc} dataset. The relative counts are counts per million tokens. Terms such as \example{attention} get a new sense increasingly used later; terms such as \example{plan} shows semantic widening moving from strong association with dialogue to other NLP tasks; terms such as \example{network} and \example{deep} show semantic narrowing moving from disperse associations to a more narrower sense associated with neural networks.}
    \label{tab:semantic-changes-s2orc}
\end{table*}
We also show an illustrative example of a change for the word \example{attention} and how it transitions according to our metric in~\autoref{fig:attention-example}.
\section{Topic Coefficients}
\label{sec:topic-coefficients}
To control for the content of the paper, we use a coarse-grained representation of the content by learning an LDA model and estimating the probability distribution of a research paper in terms of the topics. The probabilities are used as features in the regression and online prediction tasks. The regression coefficients of the topics in the full model, M4, are shown in~\autoref{tab:topic-coefficients}.
\begin{table*}[]
    \centering
    \begin{tabular}{@{}p{0.05\linewidth} p{0.1\linewidth} p{0.8\linewidth}@{}}
    \toprule

    Topic & Regression coefficients & Top words by probability\\
    \midrule
    5 & $-0.217$ & \example{user}$(0.017)$, \example{users}$(0.007)$, \example{speech}$(0.005)$, \example{knowledge}$(0.005)$, \example{generation}$(0.004)$\\ 
    6 & $-0.230$ & \example{query}$(0.016)$, \example{similarity}$(0.015)$, \example{term}$(0.014)$, \example{documents}$(0.012)$, \example{candidate}$(0.011)$ \\
    13 & $-0.064$ & \example{dialogue}$(0.030)$, \example{domain}$(0.022)$, \example{utterance}$(0.013)$, \example{dialog}$(0.012)$, \example{utterances}$(0.012)$ \\
    15 & $-0.064$ & \example{image}$(0.024)$, \example{visual}$(0.020)$, \example{object}$(0.017)$, \example{objects}$(0.016)$, \example{spatial}$(0.013)$ \\
    14 & $-0.043$ & \example{translation}$(0.057)$, \example{source} $(0.028)$, \example{target}$(0.022)$, \example{alignment}$(0.018)$, \example{parallel}$(0.013)$ \\
    12 & $-0.065$ & \example{speech}$(0.015)$, \example{chinese}$(0.015)$, \example{character}$(0.013)$, \example{languages}$(0.013)$, \example{segmentation}$(0.009)$\\
    10 & $-0.005$ & \example{lexical}$(0.010)$, \example{syntactic}$(0.010)$, \example{verbs}$(0.008)$, \example{noun}$(0.007)$, \example{argument}$(0.007)$\\
    20 & $0$ & \example{tree}$(0.021)$, \example{node}$(0.018)$, \example{nodes}$(0.013)$, \example{rule}$(0.011)$, \example{rules} $(0.011)$\\
    2 & $0.011$  & \example{classification}$(0.021)$, \example{classifier} $(0.020)$, \example{class}$(0.013)$, \example{discourse}$(0.011)$, \example{accuracy}$(0.010)$\\
    17 & $0.122$ & \example{dependency}$(0.033)$, \example{parsing} $(0.027)$, \example{parser}$(0.023)$, \example{syntactic}$(0.019)$, \example{parse}$(0.015)$\\
    3 & $0.085$  & \example{event}$(0.041)$, \example{annotation}$(0.031)$, \example{events}$(0.020)$, \example{coreference}$(0.018)$, \example{mentions}$(0.013)$\\
    9 & $0.117$ & \example{morphological}$(0.018)$, \example{pos}$(0.017)$, \example{tag}$(0.012)$, \example{tags}$(0.011)$, \example{languages}$(0.009)$\\
    11 & $0.106$ & \example{question}$(0.029)$, \example{answer}$(0.024)$, \example{questions}$(0.019)$, \example{attention}$(0.013)$, \example{dataset}$(0.012)$\\
    18 & $0.143$ & \example{sentiment}$(0.027)$, \example{tweets}$(0.013)$, \example{negative}$(0.011)$, \example{positive}$(0.011)$, \example{opinion}$(0.011)$\\
    16 & $0.149$ & \example{sense}$(0.023)$, \example{similarity}$(0.015)$, \example{wordnet}$(0.012)$, \example{senses}$(0.011)$, \example{target} $(0.009)$\\
    19 & $0.122$ & \example{topic}$(0.032)$, \example{document}$(0.026)$, \example{summary}$(0.014)$, \example{documents}$(0.014)$, \example{topics}$(0.013)$\\
    7 & $0.167$ & \example{human}$(0.006)$, \example{texts}$(0.005)$, \example{study}$(0.005)$, \example{had}$(0.004)$, \example{linguistic}$(0.004)$\\
    4 & $0.174$  & \example{relation}$(0.036)$, \example{entity}$(0.033)$, \example{relations}$(0.025)$, \example{entities}$(0.021)$, \example{knowledge}$(0.016)$\\
    1 & $0.226$  & \example{probability}$(0.016)$, \example{algorithm} $(0.010)$, \example{distribution}$(0.009)$, \example{parameters} $(0.007)$, \example{function}$(0.006)$\\
    8 & $0.334$ & \example{neural}$(0.017)$, \example{embeddings}$(0.016)$, \example{vector}$(0.013)$, \example{network}$(0.012)$, \example{embedding}$(0.012)$\\
    \bottomrule

    \end{tabular}
    \caption{\textbf{Topic coefficients and top words} We show the coefficients of the topic in the experimental model M4 alongwith the top words by probability in a given topic.}
    \label{tab:topic-coefficients}
\end{table*}
\section{Regression Results for Different Bandwidths}
\label{sec:regression-results-hyperparameters}
Different lexical and semantic influence estimates were learned by varying the bandwidth ($\gamma$). The bandwidth is a decay factor for the influence: higher bandwidth value corresponds to faster decay in influence and a lower bandwidth means a slower decay. The regressions were run for different values of the bandwidth setting \{$.001,.01,.1,1.0,10.0,100.0$\} and the optimal bandwidth was selected based on the goodness of fit on a 10\% heldout sample. The regression results for all the bandwidths are presented in~\autoref{tab:regress-citations-0001},~\autoref{tab:regress-citations-001},~\autoref{tab:regress-citations-01},~\autoref{tab:regress-citations-1} and~\autoref{tab:regress-citations-10}.

\begin{table}

\centering
\small

\begingroup
\setlength{\tabcolsep}{6pt} 
\begin{tabular}{lp{.9cm}p{.9cm}p{.9cm}p{.9cm}}
\toprule

Predictors & M1 & M2 & M3 & M4\\

\midrule

Constant&-0.000&-0.080&-0.090&-0.097\\
 &(0.005)&(0.036)&(0.036)&(0.037)\\[4pt]
Initial Citations&0.763&0.740&0.737&0.731\\
 &(0.005)&(0.005)&(0.005)&(0.005)\\[4pt]
Lex. Inf. Q2& & &0.006&0.004\\
 & & &(0.011)&(0.011)\\[4pt]
Lex. Inf. Q3& & &0.020&0.015\\
 & & &(0.014)&(0.014)\\[4pt]
Lex. Inf. Q4& & &0.066&0.054\\
 & & &(0.016)&(0.016)\\[4pt]
Sem. Inf. Q2& & & &-0.007\\
 & & & &(0.011)\\[4pt]
Sem. Inf. Q3& & & &0.018\\
 & & & &(0.014)\\[4pt]
Sem. Inf. Q4& & & &0.135\\
 & & & &(0.017)\\[4pt]
Log Lik.&-18828&-18681&-18672&-18638
\\
\bottomrule

\end{tabular}

\endgroup
\caption{\textbf{Regression analysis}. We show the results of long-term citations for various ablations when the bandwidth was set to $0.001$. The interpretation of columns and rows is similar to~\autoref{tab:regress-citations}.  
}
\label{tab:regress-citations-0001}
\end{table}
\begin{table}

\centering
\small

\begingroup
\setlength{\tabcolsep}{6pt} 
\begin{tabular}{lp{.9cm}p{.9cm}p{.9cm}p{.9cm}}
\toprule

Predictors & M1 & M2 & M3 & M4\\

\midrule

Constant&-0.000&-0.080&-0.095&-0.102\\
 &(0.005)&(0.036)&(0.036)&(0.037)\\[4pt]
Initial Citations&0.763&0.740&0.737&0.731\\
 &(0.005)&(0.005)&(0.005)&(0.005)\\[4pt]
Lex. Inf. Q2& & &0.010&0.008\\
 & & &(0.011)&(0.011)\\[4pt]
Lex. Inf. Q3& & &0.040&0.036\\
 & & &(0.014)&(0.014)\\[4pt]
Lex. Inf. Q4& & &0.060&0.048\\
 & & &(0.016)&(0.016)\\[4pt]
Sem. Inf. Q2& & & &-0.009\\
 & & & &(0.011)\\[4pt]
Sem. Inf. Q3& & & &0.021\\
 & & & &(0.014)\\[4pt]
Sem. Inf. Q4& & & &0.133\\
 & & & &(0.017)\\[4pt]
Log Lik.&-18828&-18681&-18672&-18638
\\
\bottomrule

\end{tabular}

\endgroup
\caption{\textbf{Regression analysis}. We show the results of long-term citations for various ablations when the bandwidth was set to $0.01$. The interpretation of columns and rows is similar to~\autoref{tab:regress-citations}.  
}
\label{tab:regress-citations-001}
\end{table}
\begin{table}

\centering
\small

\begingroup
\setlength{\tabcolsep}{6pt} 
\begin{tabular}{lp{.9cm}p{.9cm}p{.9cm}p{.9cm}}
\toprule

Predictors & M1 & M2 & M3 & M4\\

\midrule

Constant&-0.000&-0.080&-0.103&-0.115\\
 &(0.005)&(0.036)&(0.036)&(0.037)\\[4pt]
Initial Citations&0.763&0.740&0.736&0.729\\
 &(0.005)&(0.005)&(0.005)&(0.005)\\[4pt]
Lex. Inf. Q2& & &0.026&0.022\\
 & & &(0.011)&(0.011)\\[4pt]
Lex. Inf. Q3& & &0.036&0.028\\
 & & &(0.014)&(0.014)\\[4pt]
Lex. Inf. Q4& & &0.094&0.078\\
 & & &(0.016)&(0.016)\\[4pt]
Sem. Inf. Q2& & & &0.008\\
 & & & &(0.011)\\[4pt]
Sem. Inf. Q3& & & &0.036\\
 & & & &(0.014)\\[4pt]
Sem. Inf. Q4& & & &0.148\\
 & & & &(0.017)\\[4pt]
Log Lik.&-18828&-18681&-18663&-18625
\\
\bottomrule

\end{tabular}

\endgroup
\caption{\textbf{Regression analysis}. We show the results of long-term citations for various ablations when the bandwidth was set to $0.1$. The interpretation of columns and rows is similar to~\autoref{tab:regress-citations}.  
}
\label{tab:regress-citations-01}
\end{table}
\begin{table}

\centering
\small

\begingroup
\setlength{\tabcolsep}{6pt} 
\begin{tabular}{lp{.9cm}p{.9cm}p{.9cm}p{.9cm}}
\toprule

Predictors & M1 & M2 & M3 & M4\\

\midrule

Constant&-0.000&-0.080&-0.085&-0.106\\
 &(0.005)&(0.036)&(0.036)&(0.036)\\[4pt]
Initial Citations&0.763&0.740&0.736&0.723\\
 &(0.005)&(0.005)&(0.005)&(0.005)\\[4pt]
Lex. Inf. Q2& & &0.032&0.021\\
 & & &(0.012)&(0.012)\\[4pt]
Lex. Inf. Q3& & &0.047&0.029\\
 & & &(0.014)&(0.014)\\[4pt]
Lex. Inf. Q4& & &0.092&0.063\\
 & & &(0.017)&(0.017)\\[4pt]
Sem. Inf. Q2& & & &0.049\\
 & & & &(0.012)\\[4pt]
Sem. Inf. Q3& & & &0.097\\
 & & & &(0.015)\\[4pt]
Sem. Inf. Q4& & & &0.188\\
 & & & &(0.018)\\[4pt]
Log Lik.&-18828&-18681&-18664&-18603
\\
\bottomrule

\end{tabular}

\endgroup
\caption{\textbf{Regression analysis}. We show the results of long-term citations for various ablations when the bandwidth was set to $1.0$. The interpretation of columns and rows is similar to~\autoref{tab:regress-citations}.  
}
\label{tab:regress-citations-1}
\end{table}
\begin{table}

\centering
\small

\begingroup
\setlength{\tabcolsep}{6pt} 
\begin{tabular}{lp{.9cm}p{.9cm}p{.9cm}p{.9cm}}
\toprule

Predictors & M1 & M2 & M3 & M4\\

\midrule

Constant&-0.000&-0.080&-0.108&-0.117\\
 &(0.005)&(0.036)&(0.036)&(0.036)\\[4pt]
Initial Citations&0.763&0.740&0.727&0.718\\
 &(0.005)&(0.005)&(0.005)&(0.005)\\[4pt]
Lex. Inf. Q2& & &0.079&0.067\\
 & & &(0.012)&(0.012)\\[4pt]
Lex. Inf. Q3& & &0.097&0.075\\
 & & &(0.014)&(0.014)\\[4pt]
Lex. Inf. Q4& & &0.177&0.141\\
 & & &(0.017)&(0.018)\\[4pt]
Sem. Inf. Q2& & & &0.025\\
 & & & &(0.012)\\[4pt]
Sem. Inf. Q3& & & &0.092\\
 & & & &(0.015)\\[4pt]
Sem. Inf. Q4& & & &0.157\\
 & & & &(0.018)\\[4pt]
Log Lik.&-18828&-18681&-18615&-18569
\\
\bottomrule

\end{tabular}

\endgroup
\caption{\textbf{Regression analysis}. We show the results of long-term citations for various ablations when the bandwidth was set to $10.0$. The interpretation of columns and rows is similar to~\autoref{tab:regress-citations}.  
}
\label{tab:regress-citations-10}
\end{table}
\end{document}